\documentclass[letterpaper, 10 pt, conference]{ieeeconf}  

\usepackage{graphicx}
\usepackage[caption=false,font=footnotesize]{subfig}
\usepackage{tabu}
\usepackage{comment}
\usepackage{afterpage}
\usepackage{balance}
\usepackage{cite}
\usepackage{array}
\usepackage{makecell}

\title{Ship classification from overhead imagery \\ using synthetic data and domain adaptation}

\author{Chris M. Ward, Josh Harguess, Cameron Hilton \\
Space and Naval Warfare Systems Center, Pacific (SSC PAC) \\
San Diego, CA 92152, United States \\
Emails: \{cward, harguess, bhilton\}@spawar.navy.mil}

\begin{document}

\maketitle{}
\thispagestyle{empty}
\pagestyle{empty}

\begin{abstract}

In this paper, we revisit the problem of classifying ships (maritime vessels) detected from overhead imagery.
Despite the last decade of research on this very important and pertinent problem, it remains largely unsolved. 
One of the major issues with the detection and classification of ships and other objects in the maritime domain 
is the lack of substantial ground truth data needed to train state-of-the-art machine learning algorithms. We address this issue by building a large (200k) synthetic image dataset using the Unity gaming engine 
and 3D ship models. We demonstrate that with the use of synthetic data, classification performance increases
dramatically, particularly when there are very few annotated images used in training.

\end{abstract}

\section{Introduction}

Recently, a great deal of progress has been made in research areas within computer vision, particularly those with respect to object classification. 
This has partly been due to the advancement of deep learning approaches in convolutional neural network architectures and the availability of large annotated datasets. However, real-world applications, such as ship classification from overhead imagery, have been slow to adopt these methods due to the lack of annotated data and the difficulty of real-world image conditions such as varying lighting conditions, view angles, and atmospheric effects.

Earlier work in ship classification \cite{harguess2012vessel,rainey2011object,parameswaran2015vessel,harguess2011face} relied on classical approaches to object recognition, such as feature extraction using subspace projection and a support vector machine (SVM) \cite{scholkopf2000new} for classification. These approaches provided limited success on the practical problem of ship classification within a four-class problem (barge, cargo, container, tanker), but nevertheless underscored the importance and difficulty of the problem. 

We revisit this work with a fresh look at the problem given recent advancements, particularly in two main areas. First, recent machine learning and computer vision architectures, such as Residual Networks\cite{He2015} paired with transfer learning, have reduced the requirement of large annotated datasets in the target problem area due to their power, flexibility, and ability to learn generalized features from large datasets such as ImageNet \cite{deng2009imagenet}. Second, we build a new synthetic imagery database based on ship models taken from the same classes as the earlier work \cite{harguess2012vessel}. These synthetic images may be used to supplement the real-world image training and validation sets as shown in recent work \cite{tremblay2018training,ward2018leveraging}. In this paper, we test the ability of ResNet-34\cite{He2015} to be fine-tuned to real-world images of overhead ship imagery as well as the impact of utilizing synthetic imagery in the ship classification problem.

The paper is organized as follows: In Section \ref{methodology}, we introduce our approach to the ship classification problem as well as the baseline approaches from previous work. Section \ref{experimentation} covers the datasets created and used, the details of the experimental setup, and the experimental results. We provide a discussion on the results in Section \ref{discussion} and conclude the paper in Section \ref{conclusion}.

\section{Methodology}
\label{methodology}

We assume a typical two-stage pipeline for the overall goal of vessel classification, which includes both ship 
detection and classification. First, the ship or vessel of interest has been detected with an appropriate anomaly detection method or a ship detector trained on overhead imagery. Then, the resulting detection candidates are then fed to the second stage for classification, which is the focus of this work. 

Machine learning algorithms are trained on our synthetic image data and then fine-tuned to real-world data taken from overhead images. We use previous work in ship classification as a baseline for these algorithms. Details of the methodology are provided below.

The primary goal of our experimentation is to understand the utility of synthetic data in deep-learning-based classification of maritime vessels. We fine-tune a neural network with both synthetic imagery and real imagery and compare its performance to a variety of classic object recognition methods. We chose to use ResNet-34 as a basis for our deep learning experimentation, due to its high performance in classification tasks, as described in \cite{He2015}. While other models available at the time of our experimentation demonstrate higher classification accuracy, ResNet-34 strikes a balance between accuracy and the number of operations required for a single forward pass\cite{canziani2016analysis}. In each of our ResNet-34 experiments, we initialize the model with pre-computed ImageNet\cite{simonyan2014very,deng2009imagenet} weights. We repeated the following experiments on each of the data splits in Table \ref{table:data_splits}:

\subsection{ResNet-34 fine-tuned with BCCT200}
In order to establish baseline performance of a convolutional neural network (CNN) on BCCT200, we performed conventional transfer-learning and fine-tuning of a CNN\cite{huh2016makes}. ResNet-34 was initialized with pre-computed ImageNet weights and then end-to-end fine-tuned and tested on each of our defined BCCT200 data splits.

\subsection{ResNet-34 Fine-tuned with BCCT-Synth }
In this experiment we evaluate a CNN that is never exposed to real-world overhead imagery prior to test time; we test a ResNet-34 trained and validated only on synthetic vessel images against a set of real vessel images. We initialized ResNet-34 with pre-computed ImageNet weights and then end-to-end fine-tuned on BCCT-Synth until a validation accuracy of $99.99\%$ was achieved. The network was then tested against each defined set of BCCT200 test images.

\subsection{ResNet-34 Synthetically-Tuned with Domain Adaptation}
This experiment extends the previous experiment. We initialized ResNet-34 with our precomputed BCCT-Synth weights and then end-to-end fine-tune on BCCT200 training image sets. We refer to this final fine-tuning phase as ``domain adaptation." The domain-adapted model was then tested against each defined set of BCCT200 test images.

\subsection{Classic Object Recognition Methods}
As a baseline to the above approaches, we revisit the work in \cite{harguess2012vessel}. While there are many feature extraction and classification methods utilized in that work, we will focus our baselines in this paper on the higher performing methods. For feature extraction, Hierarchical Multi-scale Local Binary Pattern (HMLBP) \cite{guo2010hierarchical,ahonen2004face,gao2016validation}, Multi-linear Principal Components Analysis (MPCA) \cite{lu2008mpca,belhumeur1997eigenfaces}, and Histograms of Oriented Gradients (HOG) \cite{dalal2005histograms} will be used. Our baseline classifiers will be the Support Vector Machine (SVM) \cite{scholkopf2000new} and Sparse Representation-based Classification (SRC) \cite{wright2009robust}. For baseline comparisons, we will use the following combinations of feature extraction and classification methods:
\begin{itemize}
    \item HMLBP + SVM
    \item MPCA + SVM
    \item HOG +  SRC
\end{itemize}

For more details into these approaches and the parameters used for our experimentation, please see \cite{harguess2012vessel}.

\section{Experimentation}
\label{experimentation}

In this section, we introduce the datasets used in the experiments, the setup of the experiments, and the experimental results.

\subsection{Datasets}
\subsubsection{BCCT200 dataset \cite{harguess2012vessel}}
\begin{figure*}
\subfloat[Barge \label{fig:frame1}]{\includegraphics[scale=.60]{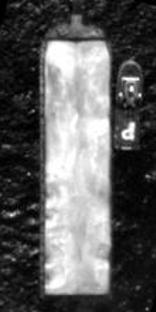}}
\hfill
\subfloat[Cargo \label{fig:frame2}]{\includegraphics[scale=.60]{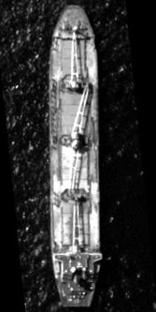}}
\hfill
\subfloat[Container \label{fig:frame3}]{\includegraphics[scale=0.60]{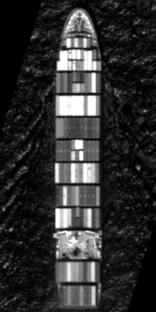}}
\hfill
\subfloat[Tanker \label{fig:frame4}]{\includegraphics[scale=0.60]{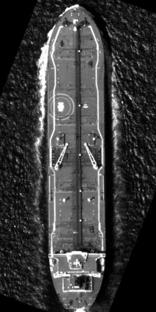}}
\hfill
\caption{Example images from the Barge, Cargo, Container, Tanker (BCCT200) dataset.}
\label{fig:bcct200} 
\end{figure*}

We will use the previously cited Barge, Cargo, Container, Tanker (BCCT200) dataset that consists of real-world overhead imagery chips of the aforementioned vessel classes, with each class consisting of $200$ images. Specifically, we utilize the BCC200-resize subset of the dataset where the images are rotated, cropped, aligned and resized. Example images of each class in BCCT200 are shown in Figure \ref{fig:bcct200}.

\subsubsection{BCCT-Synth dataset}
\begin{figure*}
\subfloat[Synthetic Barge \label{fig:synth-barge}]{\includegraphics[scale=.60]{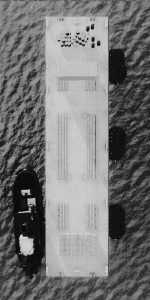}}
\hfill
\subfloat[Synthetic Cargo \label{fig:synth-cargo}]{\includegraphics[scale=.60]{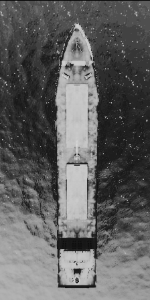}}
\hfill
\subfloat[Synthetic Container \label{fig:synth-container}]{\includegraphics[scale=0.60]{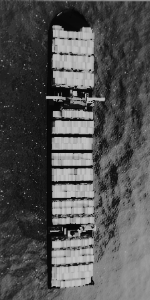}}
\hfill
\subfloat[Synthetic Tanker \label{fig:synth-tanker}]{\includegraphics[scale=0.60]{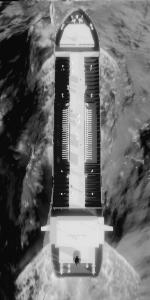}}
\hfill
\caption{Example images from the Synthetic Barge, Cargo, Container, Tanker (BCCT-Synth) dataset, which is created from 3D ship models within the Unity environment. The Synthetic Barge class features image modifiers like adjacent utility vessels and fenders (Figure \ref{fig:synth-barge}). The Synthetic Cargo, Container, and Tanker classes feature a wake modifier shown in Figure  \ref{fig:synth-tanker}. }
\label{fig:bcctsynth} 
\end{figure*}

This synthetic dataset is comprised of $200k$ labeled images of the same four generic vessel classes as BCCT200, captured from $15$ virtual overhead sensors. BCCT-Synth features imagery with varying lighting conditions, as well as varying observation angles in order to capture off-nadir instances of vessels at sea. We also introduce real-world conditions to the data, such as varying weather, sea states, and clouds. Image diversity was increased by the addition of ``class modifiers,'' including: vessel wake, secondary vessels, and fenders. Example images of each class in BCCT-Synth are shown in Figure \ref{fig:bcctsynth}.

Based on previous work in \cite{ward2018leveraging}, we used the Unity game engine as our modeling environment. After rendering, we process the Unity-generated RGB images in order to simulate the panchromatic spectrum of BCCT200. We performed histogram specification by setting the shape of the R,G,B channel histograms to match the spectral response of a panchromatic sensor. Our method of histogram matching consisted of redistribution of energy in the blue channel, and nonlinear stretch operations on red and green channels. We then performed grayscale-conversion resulting in a pseudo-panchromatic image set.

\subsection{Training and Test Data Splits}
To understand the effects of data sparsity on various object recognition approaches, we test each method against five data split ratios. The split ratios, shown in Table \ref{table:data_splits}, indicate the percentage of training images and percentage of test images used during experimentation, respectively.

Additionally, in our experiments with ResNet-34, we used a $20\%$ holdout for validation data with the exception of the 1/99 split, where a single image was used to train, and a single image was used in validation.

\begin{table}
\caption{BCCT200 training/test data split ratios and the number of images in each set. }
\begin{center}
{\small
\renewcommand{\arraystretch}{1.2}
\begin{tabu*} to 0.48\textwidth {| X[c] | X[c] | X[c] |}
\hline
Data Split \% (Training / Test) & Number of Training images & Number of Test images\\
\hline
80/20  & 160 & 40\\
\hline
50/50 & 100 & 100\\
\hline
20/80 & 40 & 160\\
\hline
5/95 & 10 & 190\\
\hline
1/99 & 2 & 198\\
\hline
\end{tabu*}
}
\end{center}
\label{table:data_splits}
\end{table}

 To assess how our models generalize to an independent data set, each object recognition method was evaluated on five data shuffles. In Table \ref{table:crossval_results} we report the average performance across the five shuffled sets. 

\subsection{Experimental Results}
\label{experimental results}
\begin{figure*}\centering
\subfloat{\includegraphics[scale=.60]{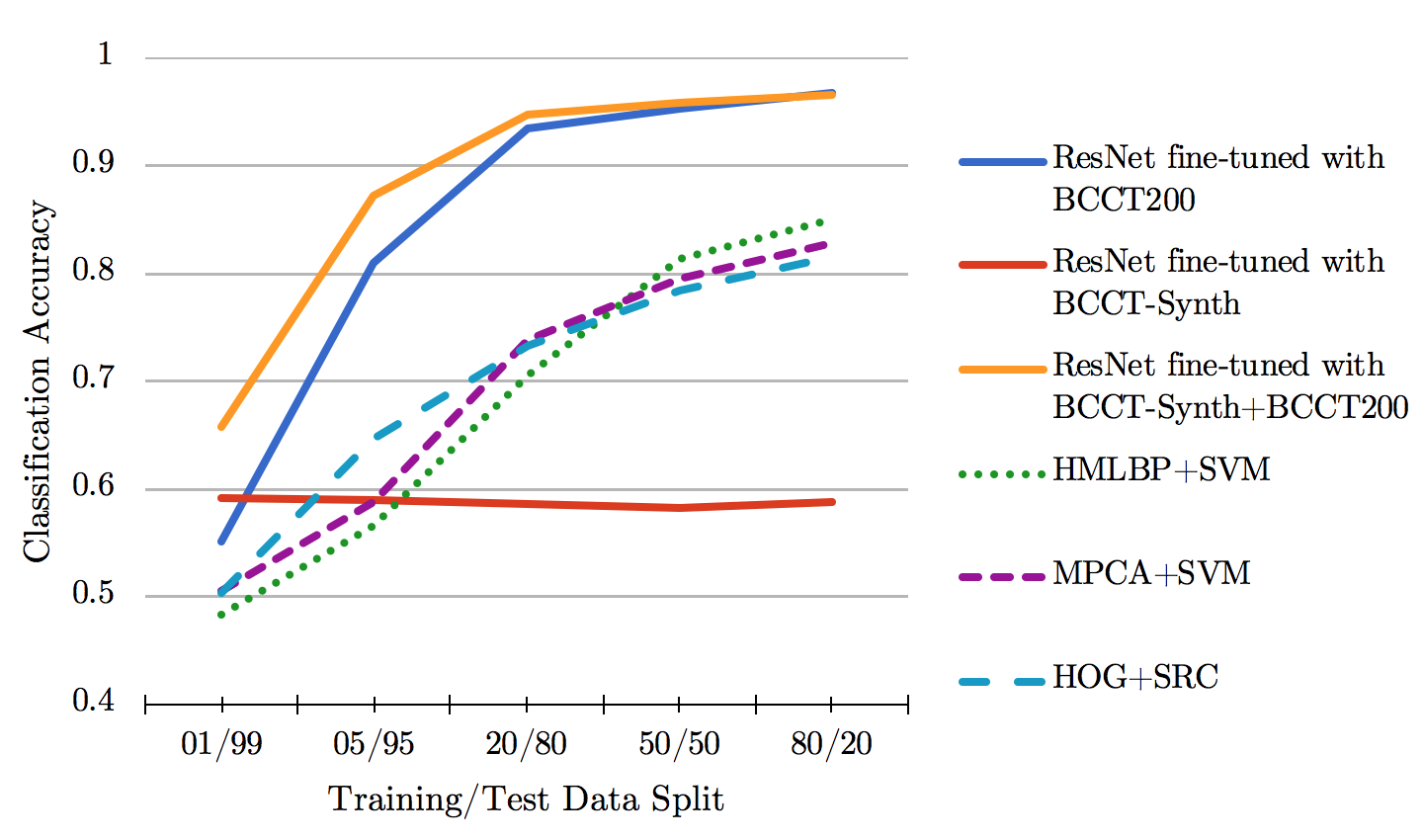}}
\caption{Classification accuracy of each tested method vs. data splits}
\label{fig:accuracy} 
\end{figure*}

The average test accuracy of 5-fold cross-validated results for each classification method under test are shown in Table \ref{table:crossval_results} and visualized in Figure \ref{fig:accuracy}. ResNet-34, fine-tuned on synthetic imagery and adapted to real imagery, outperformed other tested methods in nearly every case.

Table \ref{table:impact} shows the improvement in classification accuracy attributed to the use of synthetic imagery in the training process. With respect to the BCCT200 dataset, we observed an increasing benefit from synthetic imagery as real training data grew more sparse. As shown in Table \ref{table:impact}, synthetic imagery yielded the largest performance boost when availability of real data was minimal, yielding a $10.73$\% gain in classification accuracy.

\begin{table*}
\caption{Average of 5-fold cross-validated results for each object recognition method under test. }
\label{table:crossval_results}
\begin{center}
{\small %
\renewcommand{\arraystretch}{1.4}
\begin{tabu} to 0.98\textwidth {| X[c] | X[c] | X[c] | X[c] | X[c] |X[c] | X[c] |}
\hline
Data Split \% (Training / Test) & HMLBP+SVM & MPCA+SVM & HOG+SRC & ResNet-34 tuned on BCCT-Synth & ResNet-34 tuned on BCCT200 &  ResNet-34 tuned on BCCT-Synth + BCCT200\newline \\
\hline
80/20 & 0.8500 & 0.82875 & 0.8150 & 0.5888 & \textbf{0.9688} &  0.9663 \\
\hline
50/50 & 0.8145 & 0.7950 & 0.7840 & 0.5820 & 0.9525 &  \textbf{0.9595}  \\
\hline
20/80 & 0.7059 & 0.7375 & 0.7328 & 0.5856 & 0.9353 &  \textbf{0.9469}  \\
\hline
5/95 & 0.5653 & 0.5884 & 0.6458 & 0.5897 & 0.8092 &  \textbf{0.8730}  \\
\hline
1/99 & 0.4841 & 0.5051 & 0.5035 & 0.5920 & 0.5508 &  \textbf{0.6580}  \\
\hline
\end{tabu}
}
\end{center}
\end{table*}

\section{Discussion}
\label{discussion}

The results of training ResNet-34 on both synthetic data and varying levels of real data demonstrably suggests that synthetic data can be readily used to improve classification results of a convolutional neural network, particularly in applications where real data is extremely sparse; however, additional testing with datasets and multiple models is needed. Future experimentation on a dataset more challenging to ResNet-34 may show greater performance divergence between real and synthetically trained models.

Perhaps the greatest utility of synthetic data exists in problems where real data is not available - problem spaces where data collection is prohibitively challenging, or where object and events occur rarely and spontaneously in the wild. How might we increase the performance of a model trained on purely synthetic imagery? Recent work in computer vision implies that Generative Adversarial Networks\cite{goodfellow2014generative} have a great deal to offer as a means of domain adaptation\cite{bousmalis2017unsupervised, sankaranarayanan2017generate}, but in a scenario with extremely scarce data, or a multimodal target distribution, training an adversarial network may present a circular problem. We don't fully understand the disparity between real and synthetic data, and extending work in image quality metrics and statistical imagery analysis may provide critical insights for advancing domain adaptation techniques.

\section{Conclusion}
\label{conclusion}

Given sufficient training data, CNNs outperform classical object recognition methods. However, as training data becomes increasingly sparse, the performance of object recognition models, including CNNs, suffers. We can leverage synthetically-generated images to bolster the performance achievements of CNNs when training data is limited. The power of synthetic data paired with a deep learning model is most evident in a complete absence of real training data - our experimentation shows that a deep learning classifier trained purely on synthetic imagery can still achieve performance levels higher than traditional object recognition methods trained on real data.

\begin{table}
\caption{Impact of using synthetic imagery w/ domain adaptation}
\label{table:impact}
\begin{center}
{\small
\renewcommand{\arraystretch}{1.3}
\begin{tabu*} to 0.44\textwidth {| X[c] | X[c] |}
\hline
\makecell{Data Split \% \\ (Training / Test)} &  Accuracy Increase \\
\hline
80/20  & 0.03\%\\
\hline
50/50 & 0.70\%\\
\hline
20/80 & 1.16\%\\
\hline
5/95 & 6.34\%\\
\hline
1/99 & \textbf{10.73\%}\\
\hline
\end{tabu*}
}
\end{center}
\end{table}

\addtolength{\textheight}{-12cm}   


\bibliography{main}   
\bibliographystyle{IEEEtran}

\end{document}